\documentclass[10pt,twocolumn,letterpaper]{article}

\usepackage{cvpr}
\usepackage{times}
\usepackage{epsfig}
\usepackage{graphicx}
\usepackage{amsmath}
\usepackage{amssymb}
\usepackage{subcaption}
\usepackage[english]{babel}
\usepackage[utf8]{inputenc}
\usepackage{amsmath}  
\usepackage{algorithm}
\usepackage{algorithmic}
\usepackage{nopageno}

\usepackage[colorlinks,urlcolor=blue]{hyperref}

\cvprfinalcopy % *** Uncomment this line for the final submission

\def\cvprPaperID{13} % *** Enter the EV Paper ID here

\begin{document}

%%%%%%%%% TITLE
\title{Density Map Guided Object Detection in Aerial Images}

\author{Changlin Li$^1$, Taojiannan Yang$^1$, Sijie Zhu$^1$, Chen Chen$^1$, Shanyue Guan$^2$\\
$^1$University of North Carolina at Charlotte \quad
$^2$East Carolina University \\
{\tt\small \{cli33, tyang30, szhu3, chen.chen\}@uncc.edu, guans18@ecu.edu}
%\and
%Taojiannan Yang\\
%University of North Carolina at Charlotte\\
%9201 University City Blvd, Charlotte, NC\\
}

\maketitle
%\thispagestyle{empty}

%%%%%%%%% ABSTRACT
\begin{abstract}
   %Object detection in aerial images is a fundamental problem for a wide range of applications, \eg urban planning and traffic surveillance, and has received increasing attention recently. %However, due to the unique properties of aerial images, such as variation of viewpoint and visual distortion, general object detection algorithms fail to generate satisfactory results on aerial image datasets. 
   Object detection in high-resolution aerial images is a challenging task because of 1) the large variation in object size, and 2) non-uniform distribution of objects. A common solution is to divide the large aerial image into small (uniform) crops and then apply object detection on each small crop. 
   In this paper, we investigate the image cropping strategy to address these challenges.
   Specifically, we propose a Density-Map guided object detection Network (DMNet), which is inspired from the observation that the object density map of an image presents how objects distribute in terms of the pixel intensity of the map. As pixel intensity varies, it is able to tell whether a region has objects or not, which in turn provides guidance for cropping images statistically.
   DMNet has three key components: a density map generation module, an image cropping module and an object detector. DMNet generates a density map and learns scale information based on density intensities to form cropping regions. %Compared with the state-of-the-art ClusDet \cite{Yang_2019_ICCV}, DMNet puts more emphasis on the spatial relationship between objects. 
   Extensive experiments show that DMNet achieves state-of-the-art performance on two popular aerial image datasets, \ie VisionDrone \cite{VisionDrone} and UAVDT \cite{UAVDT}. %\textit{The source code will be publicly available.}
\end{abstract}

%%%%%%%%% BODY TEXT
\section{Introduction}
\label{sec:intro}
Object detection is a fundamental problem in computer vision, which is critical for surveillance applications, \eg, face detection and pedestrian detection. Deep learning based architectures have now become the standard pipelines for general object detection (\eg, Faster RCNN \cite{Faster_rcnn}, RetinaNet \cite{RetinaNet}, SSD \cite{SSD}). Although these methods achieve good performance on natural image datasets (\eg, MS COCO dataset \cite{MSCOCO} and Pascal VOC \cite{pascal_voc} dataset), they are not able to generate satisfactory results on specialized images, \eg, aerial and medical images. 

\begin{figure}[t]%\textcolor{yellow}{yellow}, \textcolor{orange}{orange} and \textcolor{green}{green}
\centering
\includegraphics[width=\linewidth]{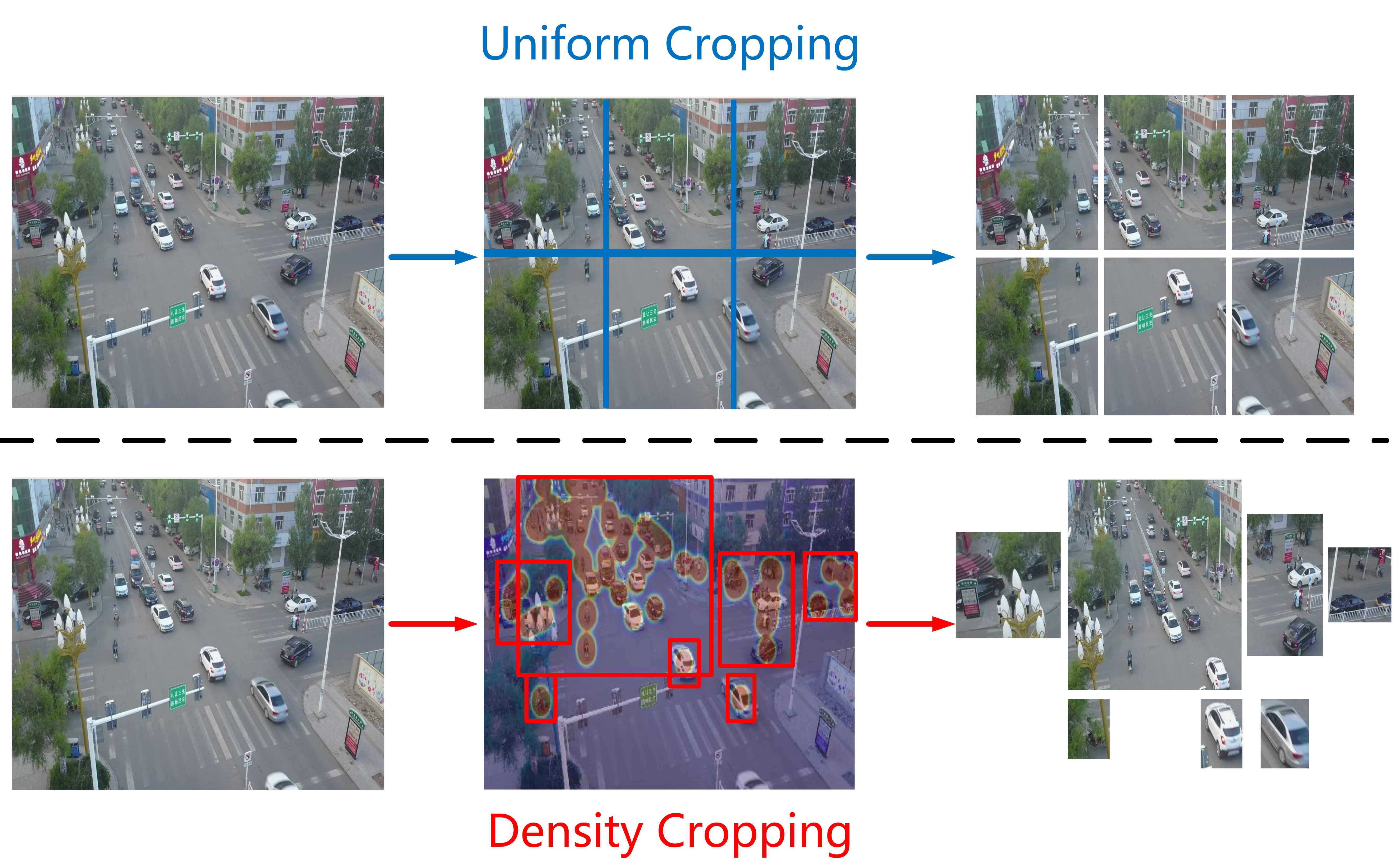}

\caption{Visualization of density cropping vs. uniform cropping. Top row provides an example of uniform cropping. Bottom row gives a comparable example of density cropping. Uniform crops have more background pixels and fail to accommodate the bounding box resolution of different categories compared with density crops. The first column shows the input aerial image. The second column shows the proposal regions for cropping. The third column shows the cropping results. Blue and red rectangles indicate candidate regions for cropping.}
\label{fig:intro}
\end{figure}

%Aerial image detection is a subset of object detection. 
Due to the special view point and large field of view, aerial image has become an important source for practical applications, \eg, surveillance. Aerial images are usually collected by drones, airplane or satellite from top view \cite{dota}, therefore their visual appearance can be significantly different from natural images like ImageNet \cite{ILSVRC15}. These characteristics give rise to several special challenges for aerial image object detection: (1) Due to variation of the photoing angle, object scale variance exists in aerial image dataset. (2) The number of objects is highly imbalanced across different categories in most of the cases. (3) Occlusion (between objects) and truncation (objects appear on the boundary) are common in aerial images. (4) Small objects account for a larger percentage compared with natural image datasets.
% and thus require more attentions.

Early works \cite{Perceptual_SOD,SOD-MTGAN} on aerial image object detection simply leverage the general object detection architecture and focus on improving the detection of small objects. \cite{Perceptual_SOD} introduces the upsampling module after feature extraction to increase spatial resolution. \cite{SOD-MTGAN} generates fine-grained feature representations to help map small objects to its larger correspondences. The improved small object detection may achieve reasonable results on popular datasets \cite{VisionDrone,VisionDrone-det,dota}, they are far from satisfactory for practical applications. 

To address the scale variation problem, another promising direction is to crop the original image into small crops/chips before applying the object detection, such as uniform cropping \cite{Unel_2019_CVPR_Workshops} and random cropping. 
%Since small object takes a higher percentage in aerial images \cite{VisionDrone} \cite{VisionDrone-det}, those methods achieve reasonable performance. 
% tial resolution \cite{Perceptual_SOD} or generate fine-grained super feature representation to help mapping small object to its large correspondent \cite{SOD-MTGAN}. Since small object takes a higher percentage in aerial image \cite{VisionDrone} \cite{VisionDrone-det}, those designs receive satisfying performance.
%On the other hand, due to the scale variance issues in aerial images, it is not straightforward for general object detectors to utilize spatial information of each category directly, since the majority of pixels are background in extreme cases. To address this issue, image cropping turns out to be a natural way to remove background. Some of the possible examples are uniform cropping \cite{Yang_2019_ICCV} and random cropping \cite{Zhang_2019_ICCV}. 
For most of the cases, these simple cropping strategies help improve the detection accuracy of small objects, since the resolutions of small crops become higher when they are resized to the size of the original image. However, they are not able to leverage the semantic information for cropping, thus resulting in a majority of crops with only background. In addition, large objects may be cut into two or more different crops by these strategies.
%the cropping strategy helps cut off background and thus increases the percentage of hard-negative examples. For detectors that require the same input size, small crops can improve the resolution of small objects by upsampling.
%can be upsampled which makes them easier to recognize.
%Those cropping methods help detect small scale objects, but it hardly benefits the overall performance. First, uniform or random cropping is easy to cut off large and medium scale objects thus is harmful to the overall performance. Second, uniform or random cropping usually results in a majority of background crops.

Following the idea of image cropping, how to find reasonable crops turn out to be critical for aerial image object detection. Apparently, cropping based on the distribution of objects would generate better crops than uniform or random strategy. And how to generate the distribution of objects has been studied in a similar task \cite{Openset-close-set}, crowd counting, which shares the same challenge of scale and viewpoint variation. In dense crowd scenes, bounding box based detection may not be applicable for small objects. Recent state-of-the-art methods leverage the power of density map for estimating the distribution of people in the scene, and achieve promising performance. This inspires us to explore the power of object density map in generating crops for aerial image object detection.
%it is hard to accurately count the number of objects in a large region, since the objects are too small and have different scales. To resolve, \cite{Openset-close-set} divides images into sub-regions, as it is easier to count in small regions compared with large regions. 
%The idea in \cite{Openset-close-set} inspires our design. To imitate similar effect in cropping, we introduce the concept of density map to aid cropping. We use a sliding window to slide on density map to estimate if an object presents in current window. In the meanwhile, we introduce a density threshold to control the size of generated region to avoid it becoming too large.  It also helps filter out regions with no objects. By merging such windows that exceeds threshold, we want to generate crops with suitable difficulty to learn.
% Similarly, we repetitively divide large aerial image into small crops. The amount of targets within each region goes down which makes it easier to learn the features and detect targets correctly.
% To improve counting accuracy, the authors divide images into sub-regions, as it is easier to count in smaller regions compared with larger regions. As the region becomes larger, the amount of features to learn increase, so does the difficulty increase. Thus it is harder to count well. To resolve the issue, by repetitively dividing feature map into small samples, the amount of target within each region goes down, and thus it is easier to learn the features and count targets correctly.

In this paper, we propose a density map based aerial image detection framework -- DMNet. It utilizes object density map to indicate the presence of objects as well as the object density within a region.
The distribution of objects enables our cropping module to generate better image crops for further object detection as shown in Fig. \ref{fig:intro}. For example, a proper density threshold can filter out most of the background area and reduce the number of objects in each crop, which makes it possible to recognize extremely small objects by upsampling the image crops. 
 % By offering a smaller amount of objects to the detector, we expect to make detection task easier to learn.

\begin{figure*}[t]
    \centering
    \includegraphics[width=\linewidth]{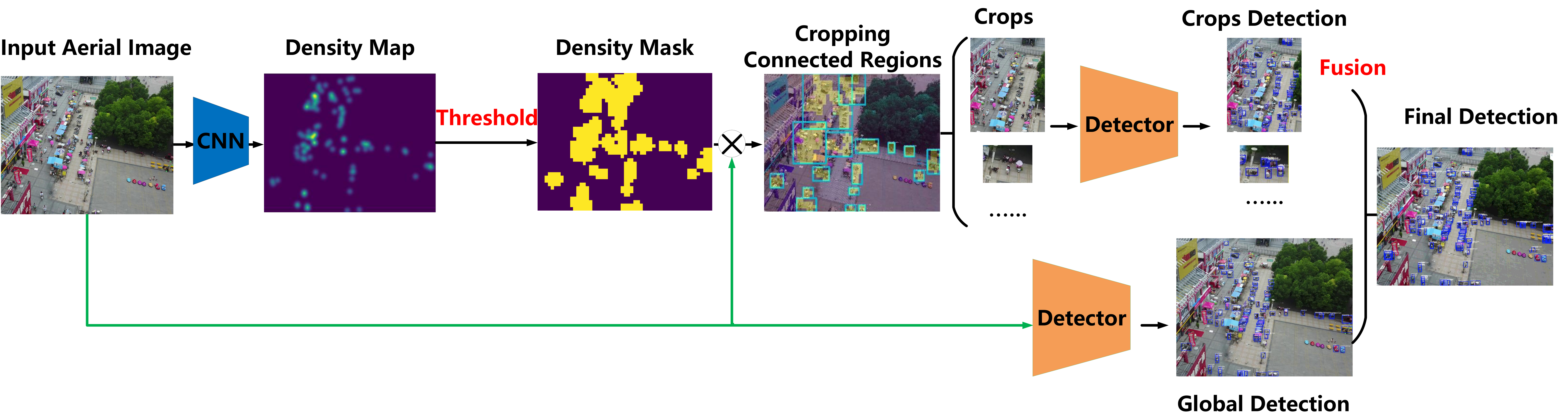}
    \caption{Overview for the DMNet framework. First, DMNet learns features of aerial images and predicts density map via the density generation module. Then it utilizes a sliding window (Section \ref{sec:density}) on the density map to obtain density mask and applies connected component algorithm to generate proposal regions for cropping. The generated image crops and the original global aerial image are feed into the same object detector for object detection.  Finally, detection results from the global image and crops are fused to generate the final detection. More details are presented in Section \ref{sec:approach}.}
    \label{fig:framework}
\end{figure*}

Fig.~\ref{fig:framework} shows the framework of the proposed DMNet. First, we introduce a density map generation network to generate the density map for each aerial image. 
% Objects are filtered by a Gaussian kernel on the density map, which provides a coarse boundary to separate objects from background pixels. 
Second, we assign a window with average object scale and slide the window over the density map without overlapping. The density map intensity indicates the probability of object presence in one position. Therefore, at each window position, the sum of all (density) pixel intensities within the window is computed, which can be considered as the likelihood of objects in this window. Then, a density threshold is applied to filter out windows with low overall intensity values. That is we assign ``0" to the window whose intensity sum value is below the threshold (\ie, the pixels in this window all have 0 value), and ``1" to the opposite.  Third, we merge the candidate windows assigned with ``1" into regions via connected component to generate image crops. Variations of pixel intensity in different regions implicitly provide the context information (\eg, background between neighboring objects) to generate valid crops accordingly. Finally, we use the cropped images to train the object detector.

Compared with existing approaches, DMNet has the following advantages: (1) It offers a simple design to crop image based on the distribution of objects with the help of object density map. (2) It is able to alleviate object truncation and preserve more contextual information than the uniform cropping strategy. (3) Compared with \cite{Yang_2019_ICCV}, which also develops a non-uniform cropping scheme, DMNet only needs to train a simple density generation network instead of training two sub-networks (\ie a cluster proposal sub-network (CPNet) and a scale estimation sub-network (ScaleNet)). 
%(4) The generated density map can also work for counting task. So it is more efficient if both detection and counting tasks are required. 

In summary, the paper has the following contributions.
\begin{itemize}
\item We are the first to introduce density map into aerial image object detection, where density map based cropping method is proposed to utilize spatial and context information between objects for improved detection performance.

\item We propose an effective algorithm to generate image crops without the need of training additional deep neural networks, as an alternative to \cite{Yang_2019_ICCV}. 

\item Extensive experiments suggest that the proposed method achieves the state-of-the-art performance on representative aerial image datasets, including VisionDrone \cite{VisionDrone} and UAVDT \cite{UAVDT}.
\end{itemize}

The rest of the paper is organized as follows. Section 2 discusses related work for object detection. Section 3 presents the methodology in detail. Section 4 provides experimental results on two datasets and extensive ablation studies. Finally, Section 5 concludes the paper.

\section{Related work}
\subsection{General object detection}
General object detection targets primarily on natural images. %As deep learning introduces stronger ability for pixel level feature extraction compared with traditional object detection methodology, lots of relevant designs have been researched and introduced for better detection performance. 
Proposal-based detectors introduce the concept of anchors with multiple stages. Fast R-CNN \cite{Fast-R-CNN} generates proposals using selective search and then extracts features and classifies objects accordingly based on those proposals. Faster R-CNN \cite{Faster_rcnn} generates proposals by the region proposal network (RPN) which significantly accelerates the inference speed. Mask R-CNN \cite{Mask-rcnn} extends Faster R-CNN to perform detection and instance segmentation tasks simultaneously. On the other hand, YOLO3 \cite{Yolo3}, SSD \cite{SSD} and RetinaNet \cite{RetinaNet} are examples of single stage detectors. Single stage detectors skip proposal stage and detect directly on sampling regions. They improve detection speed at the cost of accuracy drop. Some object detection tasks may suffer from data imbalance issue. To solve the issue, RetinaNet \cite{RetinaNet} introduces focal loss, which is a variation of cross entropy loss. It places more weights on hard examples than easy examples to guide detector to pay more attention to hard-to-learn objects.

\subsection{Object detection in aerial images}
Aerial image object detection faces more challenges compared with general object detection. First, small objects account for a higher percentage in aerial image dataset, which requires more attention on small objects \cite{VisionDrone}. Second, the object scale varies per image, per category due to the change of camera viewpoint. Third, data imbalance issue exists in aerial image dataset since some categories (such as tricycle and awning-tricycle in VisionDrone \cite{VisionDrone} dataset) rarely show in real world. Finally, aerial images may have object occlusion issue during photoing. Many research works have been developed to address these challenges.

\cite{Unel_2019_CVPR_Workshops} suggests that tiling helps improve detection performance of small objects. To counter the scale variation caused by the change of viewpoint, in \cite{Wang_2019_ICCV}, a detection network is proposed to increase the receptive field for high-level semantic features and to refine spatial information for multi-scale object detection. \cite{Yang_2019_ICCV} proposes a cluster network to crop regions of dense objects and leverages a ScaleNet to adjust generated shape of crops. The final detection result is fused from cropped images and the original image to improve overall performance. \cite{Zhang_2019_ICCV} pays attention to learn regions with low scores from a detector and gains performance by better scoring those low score regions. To solve data imbalance issue, \cite{Zhang_2019_ICCV} introduces IOU-sampling method and a balanced L1 loss. Moreover, \cite{vhr-remote-sensing, zhang2019scale} discuss challenges and insights for object detection in Very High Resolution (VHR) remote sensing imagery.

\subsection{Density map estimation}
Density map is commonly used in crowd counting literature. Crowd counting requires to estimate the head counts for a given scene where a large number of people present. Due to the high density of objects, general object detectors fail to detect and count the number of people correctly. Since density map can reflect the head location and offer spatial distribution, it turns out to be a better solution since an integral of density map can approximate head counts. Such method provides higher accuracy and thus is widely used in counting tasks. 

To improve the performance of density map based counting, \cite{MCNN} proposes geometry adaptive and fixed kernels with Gaussian convolution to generate density map. \cite{CSRNet} further improves the quality of density map by introducing a VGG16-based dilated convolutional neural network. \cite{LearnToScale} observes that the large difference in object scales leads to a great variation in density map. A scale preservation and adaption network is thus introduced to balance the pixel difference in generated density maps for robust counting performance. \cite{FastVisual} captures the pixel-level similarity in original images and implements the locally linear embedding algorithm to estimate density maps while persevering the geometry property. \cite{Examplebased} further improves the quality of generated density maps by introducing a sparsity constraint which is motivated by manifold learning.

\section{Density Map guided detection Network (DMNet)}
\label{sec:approach}
\subsection{Overview}
As shown in Fig.~\ref{fig:framework}, DMNet consists of three components, which are density map generation module, image cropping module and fusion detection module. In detail, we first train a density map generation network to predict density map for each aerial image. Afterwards, we apply a sliding window on the generated density map to gather the sum of pixel intensities and compare its value with a density threshold to form a density mask. We connect the windows whose pixel intensity is above the density threshold to generate image crops. The final detection result will be fused from detection on the image crops and the original image.

\subsection{Density map generation}
\label{sec:density}
\subsubsection{Density map generation network}
Density map is of great significance in the context of crowd counting.
%and many literatures have been devoted to relevant research. To fulfill the purpose, we adopt MCNN \cite{MCNN} in our experiments.
\cite{MCNN} proposes the Multi-column CNN (MCNN) to learn density map for crowd counting task. Due to the variation of head size per image, single column with fixed receptive field may not capture enough features. Therefore three columns are introduced to enhance feature extraction. In aerial image object detection, the general categories can be broadly divided to three sub-categories by scale (small, medium and large). 
%So we leverage a similar structure to capture the balanced feature patterns in all scales.
To capture the balanced feature patterns in all scales, we adopt MCNN \cite{MCNN} in our approach to generate object density map for image cropping.
%The backbone for density map generation modular is VGG16. We keep the first 13 layers unchanged and replaced the classification sub-network with six conv2d layers, ending with one conv1d layer to allow feature map concatenation. Note that VGG16 includes 3 pooling layers with a stride of 2 for each, resulting in the shrinking of density map resolution by the factor of 64. To preserve the original scale, we manually upsample the generated density map 64 times with cubic interpolation to restore original scale. For the case where the image height or width is not the multiplier of eight, we directly resize the image to its original scale.
% that image scales are not the multiplier of eight for height or width, those images will be resized to original scale directly to make sure the scale will match each other exactly. 

The loss function for training density map generation network is based on the pixel-wise mean absolute error, which is given as below:
\begin{equation}
L(\Theta)  = \frac{1}{2N}*\sum_{i=1}^{N}\lVert D(X_{i};\Theta)-D_{i} \rVert ^2.
\label{eq:dens-loss}
\end{equation}
$\Theta$ is the parameters of density map generation module. $N$ is the total number of images in the training set. $X_{i}$ is the input image and $D_{i}$ is the ground truth density map for image $X_{i}$. $D(X_{i};\Theta)$ stands for the generated density map by the density generation network.

%The output feature map does not have the same resolution as the input. 
As MCNN \cite{MCNN} introduces two pooling layers, the output feature map will shrink by $4 \times$ for both height and width. To preserve the original resolution, we upsample the generated density map by $4 \times$ with cubic interpolation to restore the original resolution. For the case where the image height or width is not the multiplier of four, we directly resize the image to its original resolution.

As reported in \cite{MRCNET}, it is also a working solution to add the same number of upsampling layers to restore the resolution. However, only a slight difference (approximately 0.02 in terms of mean absolute error in evaluation) is observed for this approach in our experiment. However, the size of feature maps is largely increased during training, which may cause memory issue for images with large resolution. Therefore, we do not introduce upsampling layers in our density map generation network.

\subsubsection{Ground truth object density map}
%Ground truth generation for density map remains blank in the field of aerial object detection, to the best of the authors' knowledge. To fulfill the purpose, 

To generate the ground truth object density maps for aerial images in the training stage, we follow the similar idea as proposed in \cite{MCNN} and \cite{CSRNet} for crowd counting, where two methods, geometry-adaptive and geometry-fixed kernel, are developed. 
Both methods follow the similar concepts. We use Gaussian kernel (normalized to 1 in general) to blur each object annotation to generate ground truth density maps. The key to distinguish adaptive kernel from fixed kernel is the spread parameter $\sigma$. It is fixed in fixed kernel but is computed by the $K$-Nearest-Neighbor ($K$NN) method for adaptive kernel. The formula for geometry-adaptive kernel is defined in Eq.~\ref{eq:gtg} \cite{MCNN},
\begin{equation}
F(x)  = \sum_{i=1}^{N}\delta(x-x_{i}) \times G_{\sigma_i}(x), with \  \sigma_{i} = \beta \bar{d_{i}},
\label{eq:gtg}
\end{equation}
where $x_{i}$ is the target of interest. $G_{\sigma_i}(x)$ is the Gaussian kernel, which convolves with $\delta(x-x_{i})$ to generate ground truth density map. $\bar{d_{i}}$ is the average distance of $K$ nearest targets. 
In our implementation, we prefer the fixed kernel as we consider the following assumptions for geometry-adaptive kernel are violated. (1) The objects are neither in single class nor evenly distributed per image, resulting in no guarantee for accurate estimation of geometric distortion. (2) It is not reasonable to assume the object size is related to the average distance of two neighboring objects, since objects in aerial images are not so densely distributed as in crowd counting. Based on these considerations, we choose geometry-fixed kernel accordingly.

\subsubsection{Improving ground truth with class-wise kernel}
\begin{figure}[tbp]%\textcolor{yellow}{yellow}, \textcolor{orange}{orange} and \textcolor{green}{green}
\centering
\includegraphics[width=.8\linewidth]{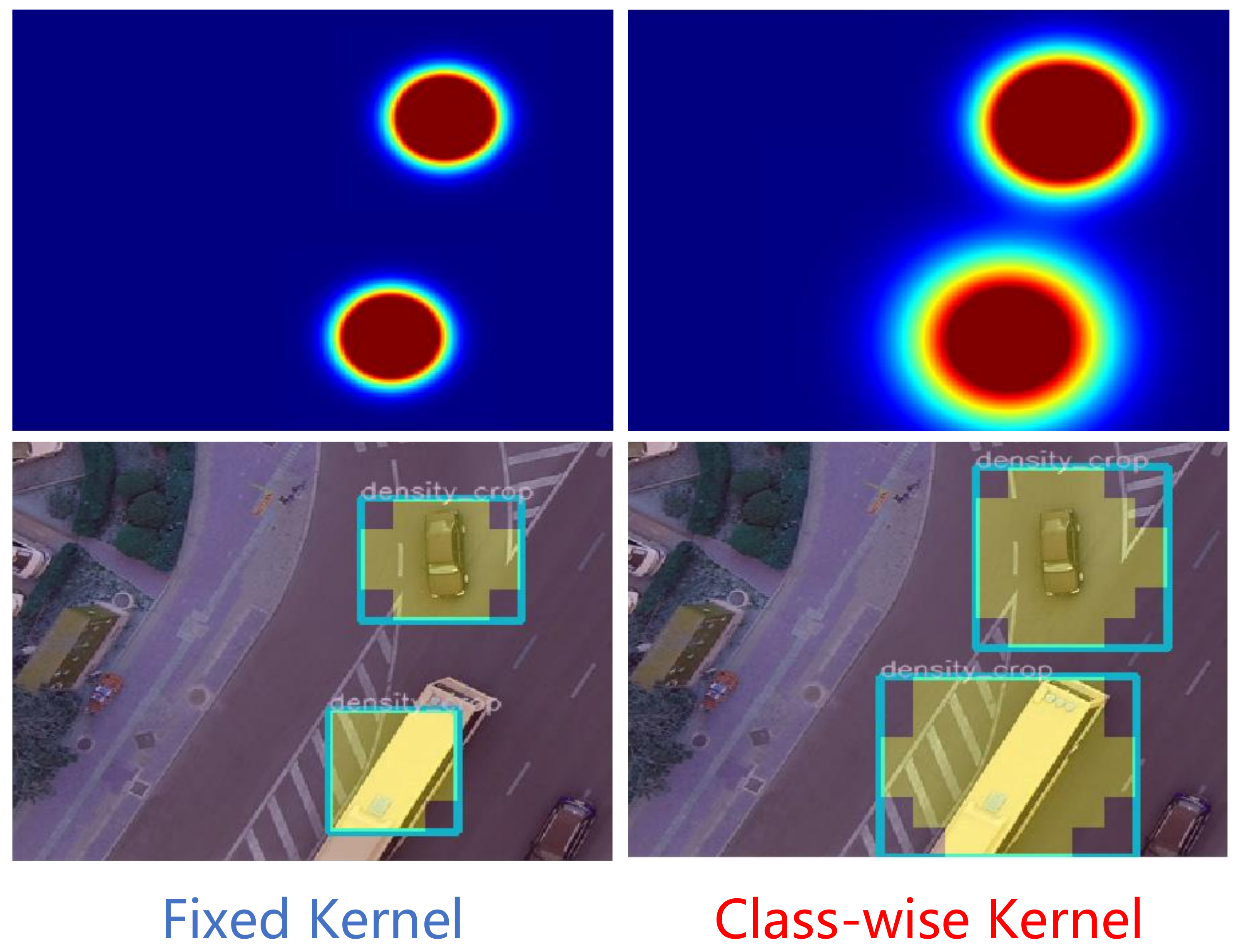}
\caption{Visual comparison between fixed kernel and class-wise kernel. Left top is the density map for fixed $\sigma$. Left bottom is its corresponding cropping results. As can be observed, the bus is not fully covered by the light blue rectangle, which results in truncation. To resolve this issue, we replace the fixed $\sigma$ with the average scale of bus category (right top). Then the light blue rectangle (right bottom) is able to fully cover the bus. Light blue rectangle represents the candidate region to crop.}
\label{fig:class-wise-kernel}
\end{figure}

% During our experiments, we observed that fixed kernel are not perfectly fit into current task either. 
In fixed kernel method, the standard deviation of Gaussian filters is constant for all objects, regardless of the shape of the exact object. This leads to possible truncation when cropping large objects (such as buses). One example is provided at the top-right of Fig. \ref{fig:class-wise-kernel}.

To resolve the possible truncation issue, we propose the class-wise density map ground truth generation method. To start, exploratory data analysis is performed on the training set to analyze the average scale for each target category. Then we compute $\sigma$ by estimating the average scale for each object category. 

Assuming that the average height and width for a category is $H_{i}$ and $W_{i}$, where $i$ is the current object category, we estimate $\sigma$ by applying Eq.~\ref{eq:class-wise-kernel}:
\begin{equation}
\sigma_{i} = \frac{1}{2}\sqrt{H_{i}^2 + W_{i}^2}.
\label{eq:class-wise-kernel}
\end{equation}

We record those $\sigma$ values for each category and apply them to Eq.~\ref{eq:gtg} to generate density maps. In this case, we are able to accommodate the scale of medium and large objects in a more suitable manner. A comparison between fixed kernel and our proposed class-wise kernel for ground truth density map generation is provided in Fig. \ref{fig:class-wise-kernel}.

\begin{figure*}[!htbp]%\textcolor{yellow}{yellow}, \textcolor{orange}{orange} and \textcolor{green}{green}
\centering
\includegraphics[width=.9\linewidth]{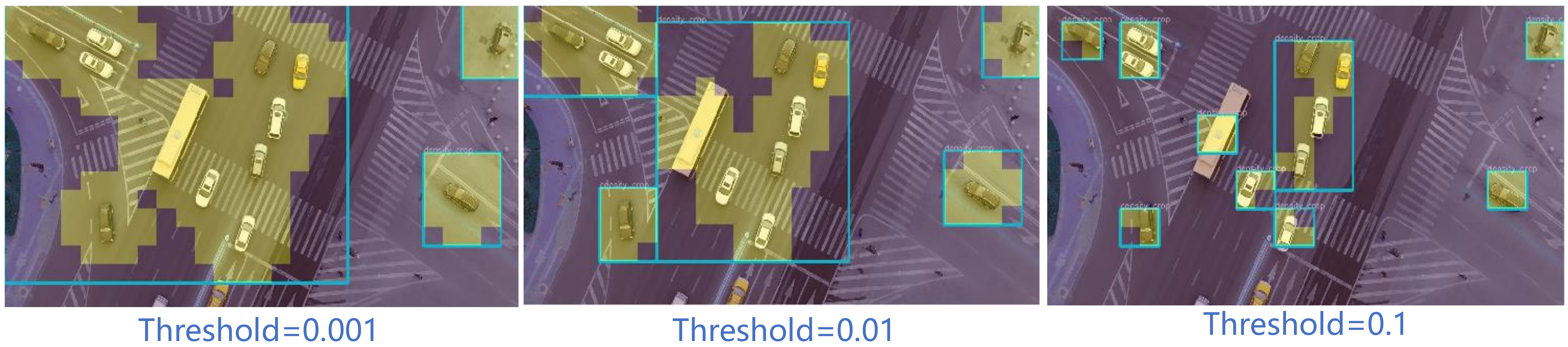}
\caption{Visualization of density mask under different thresholds. As the threshold increases, the yellow region shrinks and one large region breaks into disconnected sub-regions. 
%So more crops generated accordingly.
Yellow region is the candidate crop region and the light blue bounding box indicates the full region to crop.}
\label{fig:threshold-change}
\end{figure*}

\subsection{Image cropping based on density map}
\subsubsection{Density mask generation}

The core of DMNet is to properly crop images from the contextual information provided by density maps. As observed from the density mask provided in Fig.~\ref{fig:intro}, the regions with more objects (labeled in yellow color) have higher pixel intensities compared with those with fewer objects. By placing a threshold within a region, we can estimate the object counts and filter out pixels in the region with no or limited objects accordingly.

We introduce a sliding window on a density map, where the size of the window is the average size of the objects in the training set. We slide the window with the step of window size (\ie, non-overlapping). Then we sum all pixel values in the current window and compare the sum with the density threshold. 
%Only the pixels whose sum of intensity is greater than the density threshold will be kept. 
If the sum value is below the threshold, then the pixels in this window will all have 0 value, and ``1" for the opposite case. 
This leads to a density mask with 0 and 1 values. The detailed implementation is illustrated in Algorithm \ref{alg:dens_gen}.

The density threshold is introduced to control the noise from predicted density map. In the meanwhile, it dynamically adjusts the number of objects finally collected per density crop. By increasing the threshold, the boundary will be irregular and pixels on the boundary will be more likely to be filtered out at a higher threshold. This leads to more crops with some only have a few objects. Fig.~\ref{fig:threshold-change} provides a visualization to graphically explain how different density thresholds may affect the cropping boundary.

\begin{algorithm}
    \caption{Density mask generation}  
    \label{alg:dens_gen}  
    \begin{algorithmic}
        \STATE {\bf Input:}  Aerial image $Img$. Density map $Den$. Sliding window size $W_h, W_w$. Density threshold $TH$.
        \STATE {\bf Output:} Density mask $M$. 
        \STATE $\triangleright$ Initialization. 
        \STATE $I_h, I_w = Img.height, Img.width$. 
        \STATE $M =$ zeros $(I_h,I_w)$
        \STATE $\triangleright$ Generate density mask
        \FOR {$h$ in $range(0, I_{h}, W_{h})$}
            \FOR {$w$ in $range(0, I_{w}, W_{w})$}
            \STATE $S =$ sum $(Den[h:h + W_{h}, w:w + W_{w}])$
            \IF {$S > TH$}
                \STATE $M[h:h + W_{h}, width:width + W_{w}] = 1$
            \ENDIF
            \ENDFOR
        \ENDFOR
    \RETURN $M$
    \end{algorithmic}  
\end{algorithm}

% \begin{algorithm}
%     \caption{Implementation for density mask generation}  
%     \label{alg:dens_gen}  
%     \begin{algorithmic}
%         \STATE {\bf Input:}  $Data\_array$ Input array of tuple (Image, Density map), W$_{z}$ The average window size, W$_{th}$ The density threshold. \\
%         \STATE {\bf Output:} Density crops and associated annotations. \\
%         \FOR{$img$ and $dens$ in $Data\_list$}
%         \STATE Obtain image height $Img_{h}$ and width $Img_{w}$ \\
%         \STATE Obtain window height $W_{h}$ and width $W_{w}$ from W$_{z}$
%         \STATE Create zero mask $mask$ with the same shape of image \\ 
%             \FOR {height in $range(0, Img_{h}, W_{h})$}
%                 \FOR {width in $range(0, Img_{w}, W_{w})$}
%                     \STATE Current\_window\_threshold = dens[height:height + W$_{h}$, width:width + W$_{w}$] 
%                     \IF {$Current\_window\_threshold > w_{th}$}
%                         \STATE mask[height:height + W$_{h}$, width:width + W$_{w}$] = 1 
%                     \ENDIF
%                 \ENDFOR
            
%             \ENDFOR
%         \ENDFOR
%     \RETURN $mask$
%     \end{algorithmic}  
% \end{algorithm}

\subsubsection{Generating density crops from density mask}
The generated density mask indicates the presence of objects. We generate image crops based on the density mask. First, we select all the pixels whose corresponding density mask value is ``1''. Second, we merge the eight-neighbor connected pixels into a large candidate region. Finally, we use the candidate region's circumscribed rectangle to crop the original image. We filter out the crops whose resolution is below the density threshold. The reasons are: (1) some of the predicted density maps are not in high quality and contain noise that spreads over the whole map given a low density threshold. Thus, it is likely to obtain some random single windows as the single crop. Keeping such crops is not desired. (2) Object detectors cannot perform well on low resolution crops, as crops become really blurry after resizing to the original input size.
% To collect the density crops, we run an eight-connected-component algorithm to connect all windows that can form into regions. Then we record the left-top and right-bottom coordination of each connected region to save cropped image and annotations. Separate regions can be obtained and are picked out in a rectangle shape accordingly. We filter out the resulting bounding boxes by discarding those generated regions that fall below the minimal resolution. The reasons are, (1). Some of the predicted density maps are not in high quality and contain noise that spreads over the whole map given a low density threshold. Thus, it is likely to obtain some random single windows as the single crop. Keep such crops are not what we desired. (2). Detectors are not performed quite well for low resolution crops, as crops becomes really blur after resized to input size of detectors. 

% We present overall implementations for image cropping modular in algorithm \ref{alg:dens_gen}. It is recommended to read the algorithm for more details. Connected component algorithm has been widely applied nowadays and is easy to obtain code in different programming languages. Thus we will not discuss how to implement it in detail in pseudocode.
\subsection{Object detection on density crops}
After obtaining image crops from the density map, the next step is to detect objects and fuse results from both density crops and the whole image. Any existing modern detectors can be of the choice. We first run separate detection on original validation set and density crops. Then we collect the predicted bounding boxes from density crops detection and add them back to the detection results of original images to fuse them together. Finally, we apply non maximum suppression (NMS) to all bounding boxes and calculate the final results. The threshold of NMS is 0.5 which follows the setting in \cite{Yang_2019_ICCV}. Note that in our fusion design, we do not remove bounding boxes from original detection result. From our visualization analysis, we observe that the original detection results contain large objects that are correctly detected. Removing those detection will result in a drop in $AP_{large}$, which does not fully show the performance of the detector. Thus we keep those detected bounding boxes during evaluation. 
% However, such design may not fit each dataset. 
% We highly recommend to conduct post visualization to make decisions to fuse result for other dataset.

\begin{figure}[htp]
    \centering
    \includegraphics[width=\linewidth]{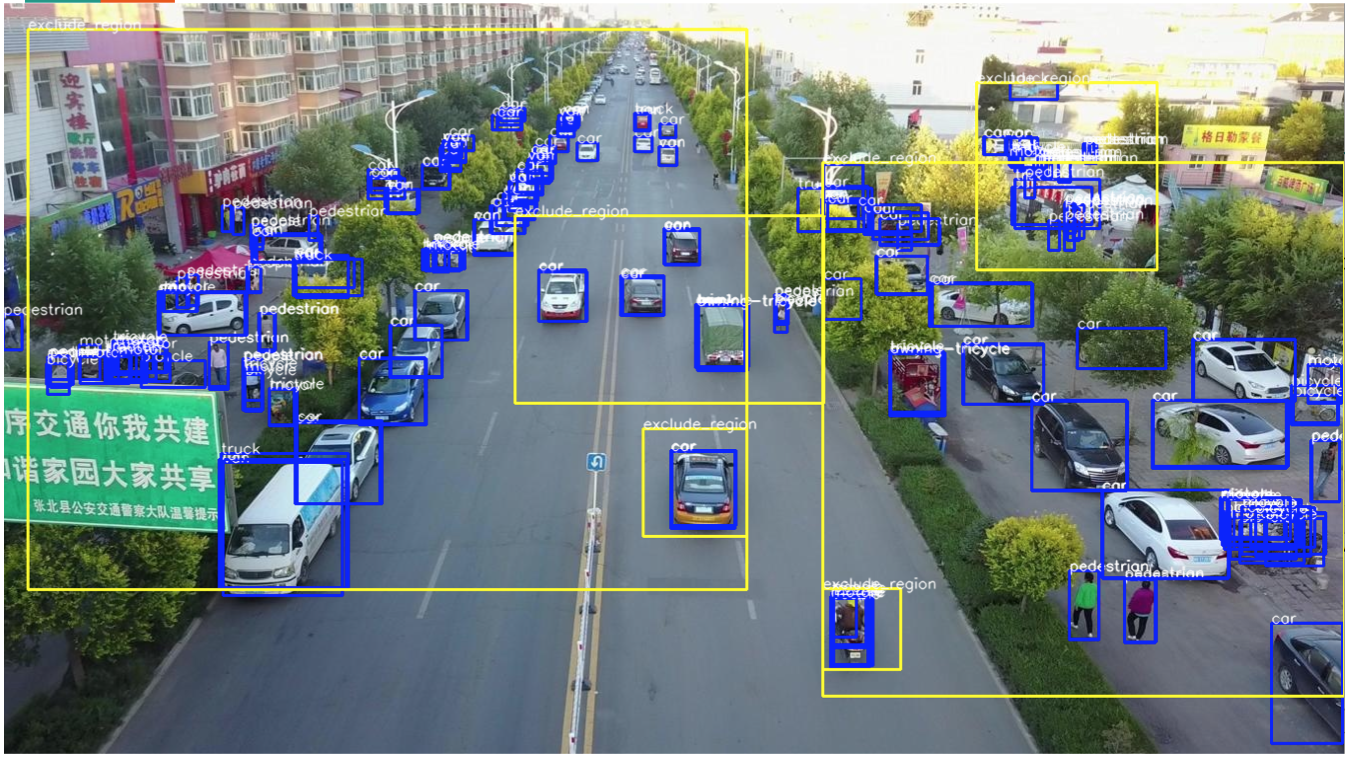}
    \caption{A visual example of the final detection result. The yellow rectangles represent regions of density crops. The blue rectangles represent ground-truth bounding boxes. The bounding boxes from both density crops and the whole images in inference stage are kept and labeled on the plot, as well as their corresponding categories. NMS is applied after obtaining the fusion bounding boxes. Thus we do not show it in this figure.}
    \label{fig:fusion}
\end{figure}

\begin{table*}[t]
\centering
\caption{Quantitative result on VisionDrone dataset. ``Test data" represents the type of data used. ``Original" is for the original validation data. ``Cluster" and ``Density" denote cluster crops \cite{Yang_2019_ICCV} and our density crops respectively. ``\#img" is the number of images that send to the detector. In the experiment, we select Average precision (AP) as the primary metric to measure the overall performance.}
\label{table:VisionDrone}
\scriptsize
\begin{tabular}{c|c|c|c|c|c|c|c|c|c}
\hline \hline
Method & Backbone & Test data &  \#Image & AP & AP$_{50}$&  AP$_{75}$ & AP$_{small}$ & AP$_{mid}$ & AP$_{large}$\\

\hline 
DetecNet+CPNet+ScaleNet \cite{Yang_2019_ICCV}   & ResNet 50   & Original+cluster    & 2716    & 26.7 & 50.6 & 24.7 & 17.6     & 38.9      & 51.4      \\
\hline 
DetecNet+CPNet+ScaleNet \cite{Yang_2019_ICCV}   & ResNet 101  & Original+cluster    & 2716    & 26.7 & 50.4 & 25.2 & 17.2     & 39.3      & 54.9      \\
\hline 
DetecNet+CPNet+ScaleNet \cite{Yang_2019_ICCV}   & ResNeXt 101 & Original+cluster    & 2716    & 28.4 & \textbf{53.2} & 26.4 & 19.1     & 40.8      & 54.4      \\

\hline \hline
DMNet & ResNet 50 & Original+density    & 2736    & 28.2 & 47.6 & 28.9 & 19.9     & 39.6      & 55.8      \\
\hline 
DMNet    & ResNet 101  & Original+density  &  2736    & 28.5 & 48.1 & 29.4 & 20.0       & 39.7      & \textbf{57.1}      \\
\hline 
DMNet                  & ResNeXt 101 & Original+density    & 2736    & \textbf{29.4} & 49.3 & \textbf{30.6} & \textbf{21.6}     & \textbf{41}        & 56.9      \\
\hline 
\end{tabular}
\end{table*}

\begin{table*}[t]
\centering
\caption{Quantitative result for UAVDT dataset.}
\label{table:UAVDT}
\scriptsize
\begin{tabular}{c|c|c|c|c|c|c|c|c}
\hline \hline
Method & Backbone  &  \#Image & AP & AP$_{50}$&  AP$_{75}$ & AP$_{small}$ & AP$_{mid}$ & AP$_{large}$\\ 
\hline
R-FCN \cite{Fast-R-CNN}     & ResNet 50 & 15096   & 7.0  & 17.5 & 3.9  & 4.4      & 14.7      & 12.1      \\ 
\hline
SSD \cite{SSD}       & N/A       & 15096   & 9.3  & 21.4 & 6.7  & 7.1      & 17.1      & 12.0      \\ 
\hline
RON \cite{Ron}      & N/A       & 15096   & 5.0  & 15.9 & 1.7  & 2.9      & 12.7      & 11.2      \\ 
\hline
FRCNN \cite{Faster_rcnn}    & VGG       & 15096   & 5.8  & 17.4 & 2.5  & 3.8      & 12.3      & 9.4       \\ 
\hline
FRCNN \cite{Faster_rcnn}+FPN \cite{FPN} & ResNet 50 & 15096   & 11.0 & 23.4 & 8.4  & 8.1      & 20.2      & 26.5      \\ 
\hline
ClusDet \cite{Yang_2019_ICCV}  & ResNet 50 & 25427   & 13.7 & 26.5 & 12.5 & 9.1      & 25.1      & 31.2      \\ 
\hline
DMNet   & ResNet 50 & 32764   & \textbf{14.7} & 24.6 & \textbf{16.3} & \textbf{9.3}      & \textbf{26.2}      & \textbf{35.2}      \\
\hline
\end{tabular}
\end{table*}

\section{Experiments}

\subsection{Implementation details}
Our implementation is based on the MMDetection toolbox \cite{chen2019mmdetection}. The MCNN \cite{MCNN} is selected as the baseline network for density map generation. For object detector, we use Faster R-CNN with Feature Pyramid Network (FPN). Unless specified, we use the default configurations for all the experiments. We use ImageNet \cite{ILSVRC15} pre-trained weights to train the detector. The density threshold is set to 0.08 in both training and testing phases for VisionDrone dataset and 0.03 for UAVDT dataset. The minimal threshold for filtering bounding boxes is set to 70 $\times$ 70, which follows the similar setting in \cite{Yang_2019_ICCV}.

The density map generation module is trained for 80 epochs using the SGD optimizer. The initial learning rate is $10^{-6}$. The momentum is 0.95 and the weight decay is 0.0005. We only use one GPU to train the density map generation network and no data argumentation is used.
% To train the model,80 epochs have been set for density generation modular with SGD optimizer. A learning rate of $10^{-7}$ has been assigned for SGD with momentum of 0.95 and the parameter decay of 0.0005. We only use single GPU to train density generation modular and no data argumentation method used for training. 

For the object detector, we set the input size to 600 $\times$ 1,000 on both datasets. We follow the similar setup in \cite{Yang_2019_ICCV} to train and test on the datasets. The detector is trained for 42 epochs on 2 GPUs, each with a batch size of 2. The initial learning rate is 0.005. We decay the learning rate by the factor of 10 at 25 and 35 epochs. The threshold for non-max suppression in fusion detection is 0.7. The maximum allowed number for bounding boxes after fusion detection is 500. Unless specified, we use MCNN to generate density map and Faster R-CNN with FPN to detect objects for all the experiments.
% We train data with 42 epochs in total, with initial learning rate of 0.005 for 2 gpus and batch size of 2. We decay learning rate by the factor of 10 after the 25 and 35 epochs. The threshold of non-max suppression in fusion detection is set to 0.7. Maximum allowed number for bounding boxes after fusion detection is set as 500. One more thing to mention, for results reported in quantitative results, we all use MCNN \cite{MCNN} and faster-RCNN \cite{Faster_rcnn} with FPN \cite{FPN} as the network backbone.

\subsection{Datasets}
To show the effectiveness of the proposed method, we evaluate the performance of DMNet on two popular datasets, VisionDrone 2018 \cite{VisionDrone} and UAVDT \cite{UAVDT}. 

\textbf{VisionDrone.} VisionDrone is a widely used dataset for aerial image detection. It includes 10,209 aerial images in total. In detail, there are 6,471 training images, 548 validation images and 3,190 testing images. Ten categories are provided for evaluation purpose with abundant annotations. The image scale is about 2,000 $\times$ 1,500 pixels. Due to the fact that we have no access to the test data and the evaluation server, we cannot evaluate our method on the test set. As an alternative, we use the validation set to evaluate the performance, which is also the choice of existing works \cite{Yang_2019_ICCV, Zhang_2019_ICCV}.

\textbf{UAVDT.} UAVDT has a rich amount of images (23,258 training images and 15,069 test images) for aerial image object detection. It has three categories, namely car, truck and bus. Those (except car) all have a larger size compared with categories in VisionDrone. The resolution for UAVDT is about 1,024 $\times$ 540 pixels.

\subsection{Evaluation metric}
We follow the same evaluation metric as proposed in MS COCO \cite{MSCOCO}. Six evaluation metrics are employed, namely AP (average precision), AP$_{50}$, AP$_{75}$, AP$_{small}$, AP$_{medium}$ and AP$_{large}$. The AP is the average precision under multiple IoU thresholds, ranging from 0.50 to 0.95 with a step size of 0.05. Since AP considers all thresholds, we use AP to measure and compare the performance between the proposed method and other competing approaches. Meanwhile, as the number of generated image crops will affect the inference speed, we also record image counts in the table for a fair comparison. We denote ``\#img" for the total number of images (including both original images and density crops) we used in the validation set.

\subsection{Quantitative result}
In this section, we evaluate the proposed DMNet on VisionDrone and UVADT datasets. Table \ref{table:VisionDrone} shows the results on VisionDrone. We can see that DMNet consistently outperforms ClusDet \cite{Yang_2019_ICCV} by 1-2 points on three different backbone networks. Specifically, DMNet achieves the state-of-the-art performance of 29.4 AP with the ResNetXt101 backbone. This clearly exceeds all previous methods. Moreover, the result of AP$_{75}$ improves nearly 4 points compared with ClusDet \cite{Yang_2019_ICCV}, indicating the robustness of DMNet at higher IoU thresholds. We also observe more than 2 points improvements on AP$_{small}$ under different backbones, which suggests that the proposed density map crops significantly help the detection for small scale objects. 

% Table~\ref{table:VisionDrone} shows the comparison result with benchmarks from \cite{Yang_2019_ICCV}. To be fair, we only compare the result from the same backbone. Our model achieves the best of 29.4 AP with the backbone of ResNetXt101, which exceeds the performance of all other benchmarks. In the meanwhile, the result of AP 75 improves nearly 4 points compared with ClusDet, indicating the robustness of DMNet at higher IOU threshold. We also observed 2 points improvements for AP$_{small}$, which suggests the stronger detection ability for small objects. 

Table~\ref{table:UAVDT} shows the results of different methods on UVADT. It can be seen that general object detectors fail to achieve a comparable result as discussed in Sec \ref{sec:intro}. Similar to the results in VisionDrone, DMNet substantially outperforms ClusDet and achieves the state-of-the-art performance of 14.7 AP on UVADT. Particularly, DMNet consistently improves the accuracy on small scale, medium scale and large scale objects. This validate the effectiveness of our generated crops based on density maps.
% comparison result with benchmarks from \cite{Yang_2019_ICCV}. DMNet obtains the highest AP amongst all available benchmarks with the best AP of 14.7, which further improves 1 point compare with ClusDet. We get similar AP$_{small}$ but is able to further improve the rest category with average of 2 AP. This clearly indicates the stronger detection performance of DMNet. 

\textbf{Inference speed.} Here we report the inference speed of the proposed DMNet. We conduct the experiment on one GTX 1080 Ti GPU per task. The inference speed on three backbones (ResNet 50, ResNet 101 and ResNeXt 101) is 0.29 s/img, 0.36 s/img and 0.61 s/img, respectively.

\begin{table*}
\centering
\caption{Ablation study on VisionDrone Dataset.}
\label{table:ablation}
\begin{tabular}{|c|c|c|c|c|} 
\hline
Method  & AP & AP$_{small}$ & AP$_{mid}$ & AP$_{large}$    \\  
\hline
FRCNN \cite{Faster_rcnn}+FPN \cite{FPN} & 21.4 & 11.7 & 33.9 & 54.7\\  
\hline
DMNet without thresholding & 22.6  & 11.8 & 37.5 &58.5\\ 
\hline
Uniform cropping without fusion& 24.5 &19.1  & 31.9 & 22.4\\
\hline
DMNet without fusion &25.9 &19.4 &38.1 &41.6 \\
\hline
DMNet with all components  & 28.2 & 19.9 & 39.6 &55.8\\
\hline
\end{tabular}
\end{table*}

\begin{figure*}[t]
    \centering
    \includegraphics[width=\linewidth]{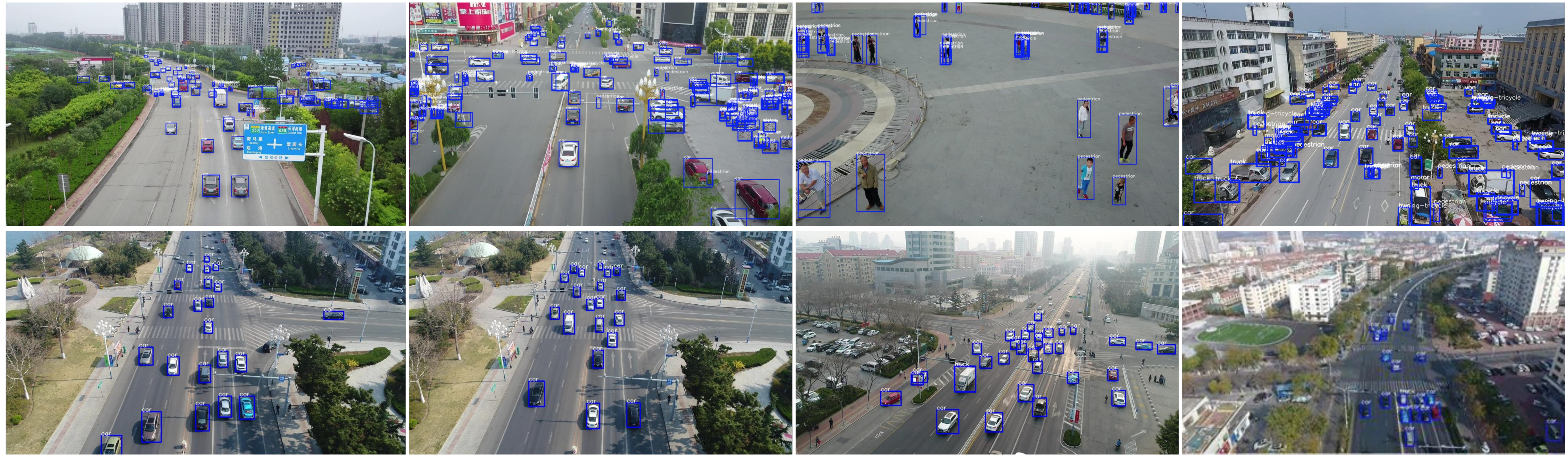}
    \caption{Visualization of our DMNet detection results on VisionDrone (first row) and UAVDT (second row). }
    \label{fig:final_det_result}
\end{figure*}

\subsection{Ablation study}
In this section, we design a series of ablation studies to analyze the contribution of each component in the proposed DMNet. 
% experiment to check how will the performance of DMNet changes without some components in out experiment. 
In all experiments, we use MCNN \cite{MCNN} as the density generation backbone and Faster RCNN \cite{Faster_rcnn} as the detector. The input image size is 600 $\times$ 1000.

\textbf{Density threshold.} The density threshold is an important factor as it controls how to generate density crops. In this experiment, we remove thresholding by keeping all windows whose pixel intensity is larger than 0. From Table~\ref{table:ablation} we can clearly see that AP drops drastically without thresholding. From the previous result analysis, we examine the generated crops and find most of them are large and cover many objects, which makes it difficult to detect small objects. Since no threshold is applied, more background pixels are cropped, which further affect the performance of detector. 
%So we eventually observe large drop for AP score.

%\textbf{Density map generation network.} Our images crops is generated from the predicted density map. Since a high quality density map can help generate better crops, it may also improve final detection performance. 
%To see if this is true, we switch the density generation modular from MCNN \cite{MCNN} to CSRNet \cite{CSRNet}. As thresholding may affect generated crops, we set three different thresholds,0.01.

%Table provides details. Both receives similar detection performance so changes of backbone may not affect much. We also conduct experiment for no thresholding situation, where CSRNet \cite{CSRNet} (AP 23.8) get a better performance compared with MCNN \cite{MCNN} (AP 22.6).

\textbf{Comparison with uniform crops.} As discussed in Sec \ref{sec:intro}, aerial images contain a majority of small scale objects. DMNet is able to effectively crop small objects from the whole image and significantly improve AP$_{small}$ as stated in Table \ref{table:VisionDrone}. But one can also get small objects by uniform cropping with a very small window size. In this experiment, we replace our density crops with $3 \times 4$ uniform cropping, where the size of each uniform crop is small to benefit small object detection. As shown in Table~\ref{table:ablation}, this method fails to beat DMNet, although it improves nearly 3 points on AP compared with the baseline. The reason is that although small uniform crops are able to help small object detection, they also increase the risk of cutting off large objects. We can see that the AP$_{small}$ is comparable with DMNet while there is a large drop in AP$_{medium}$ and AP$_{large}$. This demonstrate the superiority of our DMNet since it is able to better accommodate object scales and thus achieves better performance.

% As a higher threshold selected, the generated crops tend to have less object. Since uniform crops with tiny steps can achieve similar effect, we want to compare their performance. To fulfill the objective, we double uniform crops from 6 to 12. As the result from table~\ref{table:ablation}, it fails to beat DMNet, although it improves nearly 2 points compared with 6 crops. From post analysis, we find the reason, which is due to the large drop in AP$_{medium}$ and AP$_{large}$ category. This in turn suggests the robostness of DMNet as it is able to adaptively recognize the object scale and thus achieve better performance.

\textbf{Contribution of density crop detection}. Directly detecting objects on image crops instead of the original image can give better performance as reported in \cite{Yang_2019_ICCV}. However, how it contributes to the final fusion detection remains unclear. Therefore, we additionally report performance of DMNet with only detection on images crops (\ie, without fusing the results of detection on the original whole images).
The results are provided in Table~\ref{table:ablation}. We can conclude that density crop detection primarily contributes to AP$_{small}$ and AP$_{mid}$ as the large performance improvements have been observed on those two categories. Meanwhile, detection on the original image contributes more on the AP$_{large}$ category, compared with density crop detection. 
%So to summarize, density crop contributes more to AP$_{small}$ and AP$_{mid}$, while detection on original image contributes to AP$_{large}$ respectively. 
% Detailed statistics provided in Table~\ref{table:ablation}.

\section{Conclusion}
In this paper, we propose the density map guided detection network (DMNet) to address the challenges in aerial image object detection. Density map provides spatial distribution and collects window-based pixel intensity to implicitly form the boundary of a potential cropping region, which benefits the following image cropping process. The proposed DMNet achieves state-of-the-art performance on two popular aerial image detection datasets under different backbone networks. Extensive ablation studies are conducted to analyze the contribution of each component in DMNet. Our proposed density map based image cropping strategy provides a promising direction to improve the detection accuracy in high resolution aerial images.
% rove the evaluation performance of aerial image detection. We show that by introducing density map, it helps utilize spatial distribution and is able to reflect the possible boundary of a potential cropping region, which in turn guiding image cropping process. We explore the possible factors that could further improve the performance of DMNet. The quantitative results on VisionDrone \cite{VisionDrone} and UAVDT \cite{UAVDT} suggest the effectiveness of our method.
%------------------------------------------------------------------------

\section{Acknowledgements} 
This material is based upon work supported by the National Science Foundation under Grant CNS-1910844. Any opinions, findings, and conclusions or recommendations expressed in this material are those of the author(s) and do not necessarily reflect the views of the National Science Foundation.

{\small
\bibliographystyle{ieee_fullname}
\bibliography{egbib}
}

\end{document}